\documentclass[conference]{IEEEtran}
\IEEEoverridecommandlockouts
\usepackage{cite}
\usepackage{amsmath,amssymb,amsfonts}
\usepackage{algorithmic}
\usepackage{graphicx}
\usepackage{textcomp}
\usepackage{xcolor}
\usepackage{subfigure}
\usepackage{color}
\usepackage{graphicx}
\usepackage{booktabs}
\usepackage[misc]{ifsym}

\newcommand{\gray}[1]{\textcolor{gray}{#1}}
\newcommand{\green}[1]{\textcolor[RGB]{96,177,87}{#1}}
\newcommand{\fn}[1]{\footnotesize{#1}}

\newcommand{\gbf}[1]{\green{{\fn{ (#1)}}}}
\newcommand{\rbf}[1]{\gray{{\fn{ (#1)}}}}
\newcommand{\rrbf}[1]{\gray{{\fn{#1}}}}

\def\BibTeX{{\rm B\kern-.05em{\sc i\kern-.025em b}\kern-.08em
    T\kern-.1667em\lower.7ex\hbox{E}\kern-.125emX}}

\begin{document}
\author{\IEEEauthorblockN{Anonymous Authors}}

\title{Feature Augmentation for Self-supervised Contrastive Learning: A Closer Look
\thanks{\Letter \space indicates orresponding author. This work is supported by Shenzhen Science and Technology Innovation Commission (JCYJ20200109114835623, JSGG20220831105002004), and Guangdong Provincial Key Laboratory of Cloud Security Key Technology (2022B1212020006).}
}

\author{
    Yong Zhang\textsuperscript{\dag, \S}, Rui Zhu\textsuperscript{\ddag}, Shifeng Zhang\textsuperscript{\S}, Xu Zhou\textsuperscript{\S}, Shifeng Chen\textsuperscript{\dag}, and Xiaofan Chen\textsuperscript{\S, \Letter}\space\\
    \small{\textsuperscript{\dag}Shenzhen Institute of Advanced Technology, CAS \space\space \textsuperscript{\ddag}The Chinese University of Hong Kong, Shenzhen \space\space \textsuperscript{\S}Sangfor Technologies Inc.}\\
    {\tt\scriptsize\{yongzhang,ruizhu\}@link.cuhk.edu.cn, \{zhangshifeng,zhouxu,chenxiaofan\}@sangfor.com.cn, shifeng.chen@siat.ac.cn}
}


\maketitle

\begin{abstract}
Self-supervised contrastive learning heavily relies on the view variance brought by data augmentation, so that it can learn a view-invariant pre-trained representation. Beyond increasing the view variance for contrast, this work focuses on improving the diversity of training data, to improve the generalization and robustness of the pre-trained models.
To this end, we propose a unified framework to conduct data augmentation in the feature space, known as feature augmentation. This strategy is domain-agnostic, which augments similar features to the original ones and thus improves the data diversity. We perform a systematic investigation of various feature augmentation architectures, the gradient-flow skill, and the relationship between feature augmentation and traditional data augmentation. Our study reveals some practical principles for feature augmentation in self-contrastive learning. By integrating feature augmentation on the instance discrimination or the instance similarity paradigm, we consistently improve the performance of pre-trained feature learning and gain better generalization over the downstream image classification and object detection task.
\end{abstract}


\section{Introduction} \label{sec:intro}

\begin{figure}[tb!]
    \begin{center}
        \includegraphics[width=0.95\linewidth]{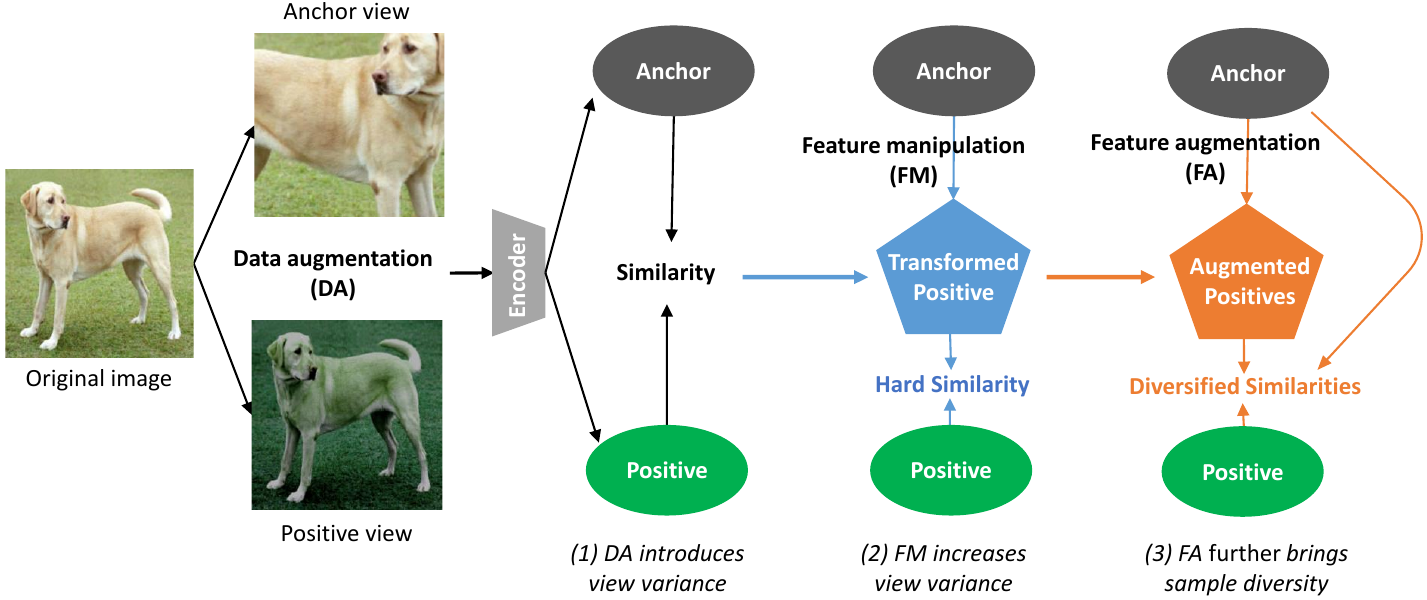}
    \end{center}
    \caption{In self-supervised contrastive learning, data augmentation (DA) introduces view variance usually by transforming the original input. Instead, feature augmentation (FA) aims to further increase the sample diversity in the feature space after the encoder. Feature manipulation (FM) is a special case of FA that focuses on mining hard examples.}
    \label{fig:dafa}
    \vspace{-0.6cm}
\end{figure}

Data augmentation (DA) has become a widely employed regularization strategy for model generalization in supervised learning, which is essential especially when encountering insufficient training data. Though numerous unlabeled data are usually available in the self-supervised learning setting, data augmentation also plays an important role in some self-supervised contrastive learning paradigms \cite{caron2020unsupervised,chen2020simple,tian2020makes,grill2020bootstrap,van2021revisiting,cole2022does}. For example, as shown in Fig. \ref{fig:dafa}, given an image, data augmentation is leveraged to generate various views (i.e., a particular feature group in multi-view learning \cite{xu2013survey}) to form contrastive pairs for feature learning.
By maximizing the mutual information among the augmented views from one instance (i.e., positive pairs), the contrastive pre-trained representation can even achieve comparable transfer performance with its supervised counterpart \cite{hjelm2018learning}. 

The success of self-supervised contrastive feature learning relies on how data augmentation is structured. SimCLR \cite{chen2020simple} and BYOL \cite{grill2020bootstrap} have explored various data augmentation strategies by introducing some image transformations to increase the view variance. However, these data augmentation strategies have several intrinsic limitations:   
(1) Data augmentation is generally domain-specific. For example, colorization may work effectively for images but not for videos. Hence, designing an effective DA approach requires experienced domain knowledge\cite{lee2020mix,verma2021towards}. 
(2) Data augmentation incurs task bias. Previous works \cite{wang2020dense,xie2021detco,xie2020propagate,zhao2020makes} have shown that pre-trained models can transfer well to the downstream task of image classification rather than object detection. 
(3) Data augmentation is inflexible. For instance, data augmentation via geometric transformations (e.g., cropping, color distortion) typically needs a grid search to find the optimal parameter setting. Previous works \cite{tian2020makes,chen2020simple,grill2020bootstrap} have to conduct a customized search to find the sweet-spot parameter of an augmentation.

To tackle the aforementioned limitations, feature augmentation (FA) \cite{devries2017dataset} is introduced to conduct ``data augmentation'' in the feature space instead of the original input space. This means we can search feature points to enlarge training data. Notably, since FA does not directly work on the raw data, it is a domain-agnostic augmentation. Secondly, FA can better tackle the task-bias problem thanks to fewer hand-designed operations. Thirdly, as we see from Fig. \ref{fig:dafa}, FA is performed by merely manipulating the encoder's feature vectors, which makes FA very flexible.

Feature augmentation for self-supervised learning \cite{balestriero2023cookbook} is an underexplored problem so far. Previous works \cite{verma2019manifoldmix,zhu2021improving,kalantidis2020hard_mochi} propose to adapt the distribution of training data by transforming the original data in the feature space. This is often referred to as feature manipulation (FM), including methods like manifold-mixup \cite{verma2019manifoldmix,zhu2021improving} and nearest-neighbor operation \cite{dwibedi2021little,kalantidis2020hard_mochi}. 
Basically, FM can be regarded as a special case of FA. As shown in Fig. \ref{fig:dafa}, except for searching hard examples in FM, FA further pursues the diversity of samples to obtain better robustness and generalization. To achieve general FA, there are still three open problems:
(1) \textit{Architecture}: which network architecture is suitable for FA? SimCLR \cite{chen2020simple} and BYOL \cite{grill2020bootstrap} utilize a feature projector and a predictor, while MoCoV1 \cite{he2020momentum} only has a light MLP head. How to use the projector and predictor and where to perform FA need further investigations.
(2) \textit{Gradient flow}: how to control the flow of gradient back-propagation? The gradients in DA propagate through the main encoder, and then the model learns the view variance from DA. Unlike DA conducted at the input level, FA performs feature-level augmentation. Hence, FA cannot propagate through the main encoder, which makes the model easier to overfit. Therefore, it is essential to control the gradient flow to avoid overfitting.
(3) \textit{The relationship between FA and DA}: the nearest neighbors of a sample instance can serve as a proxy for strong DA \cite{koohpayegani2021mean}. However, the relationship between FA and DA is still unclear. What kind of DA setting is good for FA? Could FA make up a deficiency when FA is insufficient?

To address these problems, we have made a systematic study on FA for self-supervised contrastive learning. Concretely, we integrate both the instance discrimination (cross-entropy-based, e.g., \cite{chen2020simple}) and the instance similarity (prediction-based, e.g., \cite{grill2020bootstrap}) self-supervised contrastive learning paradigms \cite{balestriero2023cookbook} into a unified framework. Then, we thoroughly investigate the network architecture for optimal learning performance. We conduct an empirical evaluation of several proposed FA methods to verify their effectiveness. During this study, we also find out some practical principles for effectively applying FA in self-supervised contrastive pre-training.

In summary, we have made the following contributions:
\begin{itemize}
	\item We propose a unified FA framework for self-supervised contrastive learning, which enables us to explore various network layouts of FA to achieve optimal performance.
	\item We conduct a systematic study on FA from three perspectives: the network architecture, the gradient flow, and the relationship between FA with DA.
	\item Our proposed FA framework brings consistent performance improvements on the downstream image classification task as well as better generalization for object detection, by integrating the instance discrimination and the instance similarity pre-training paradigms.
\end{itemize}

\section{Related Work}

{\textbf{Data augmentation for contrastive learning.}}
Data augmentation (DA) \cite{yang2022image} is the most indispensable part of self-supervised contrastive learning as mentioned in \cite{le2020contrastive}. For example, in terms of DA's difficulty: SimCLR \cite{chen2020simple} investigates the importance of DA and finds out that contrastive pre-training needs stronger DA than the supervised counterpart. 
Subsequently, InfoMin \cite{tian2020makes} suggests increasing DA to the ``sweet spot'' of mutual information can improve transfer performance.
In terms of the number of augmented views, SimCLR \cite{chen2020simple}, MoCo \cite{he2020momentum}, BYOL \cite{grill2020bootstrap} and most contrastive methods\cite{bachman2019learning,tian2020makes,zhao2020makes,chuang2020debiased} apply two views to construct one positive pair. 
Beyond two views, SwAv \cite{caron2020unsupervised} proposes multi-crop to augment additional small views to construct multiple positive pairs.
Besides, ReSSL \cite{zheng2021ressl} and MSF \cite{koohpayegani2021mean} discuss their superiority based on weak/strong data augmentation settings.

{\textbf{Mixup based contrastive pre-training.}}
Mixup \cite{zhang2017mixup} offers a solution for domain-agnostic pre-train. By simply interpolating two data points, mixup could be a general DA strategy without the specific even laborious design for a certain modality or a particular domain (e.g., color jittering for natural image and word mask for language).  
I-Mix \cite{lee2021imix} and DACL \cite{verma2021towards} verify the efficacy of Mixup in both vision and non-vision modalities. 
In the medical image domain, C2L \cite{zhou2020comparing} proposes a large-scale self-supervised contrastive model and shows the superiority of Mixup.
For the natural image domain, Un-mix\cite{shen2020mix} empirically reveals the flexibility, universality and consistent performance improvement of mixup on a variety of main-stream self-supervised methods.

{\textbf{Feature manipulation in self-supervised learning.}}
We have mentioned the application of feature manipulation on visual self-contrastive leaning\cite{kalantidis2020hard_mochi,zhu2021improving,dwibedi2021little,koohpayegani2021mean}. 
Besides, Wu et.al \cite{wu2023hallucination} propose feature hallucination which focuses on augmenting an extra positive view in the feature space for self-supervised contrastive learning. Metaug \cite{li2022metaug} performs a meta-learning strategy to construct the augmentation generator updated by the reward from the encoder and directly augment the discriminative features in the feature space. From the new perspective for energy-based contrastive learning \cite{kim2022energy}, VEM \cite{du2022variational} generates negative features from the energy-based feature distribution in self-supervised contrastive learning. 
More recently, I-JEPA~\cite{assran2023self} proposed to perform prediction in the feature space of the joint-embedding architecture, which is somewhat similar to the philosophy of such a feature-level augmentation method.

{\textbf{Feature augmentation beyond self-supervised learning.}}
The idea of augmenting samples in feature space has been explored for several years. 
DeVries et.al \cite{devries2017dataset} first investigate FA as a domain-independent and general-purpose strategy to improve generalization with limited labeled samples in various modalities (e.g., speech, sensor processing, motion capture, and images). 
In the NLP community, Kumar et.al \cite{kumar2019closer} compare six FA methods on the few-shot intent classification task and FA provides an effective way to improve the performance with very few available samples.
Similarly in low-shot visual learning, Hariharan et.al \cite{hariharan2017low} present a FA method for hallucinating additional examples for the data-starved categories by transferring variation from the base categories.
In supervised metric learning to improve the performance, the embedding expansion technique \cite{embed_expan} selects the hardest negative pair from the interpolation between positives and negatives. DVML\cite{dvml} and DAML\cite{daml} generate new hard examples by variational and adversarial generators. 

\section{Feature Augmentation for Contrastive Learning: Approach \& Exploration Experiments}

In this section, we first represent the basic contrastive learning framework with FA. Next, we illustrate the adopted FA methods for contrastive learning. 
After that, we empirically evaluate the three unsolved problems: 
1) we leverage the projector and predictor modules to equip with FA and select the proper architecture;
2) we validate the significance of the stop-gradient for the gradient-flow problem;
3) we empirically explore the relationship between FA and DA, based on which we confirm the ultimate parameter setting of data augmentation.
Finally, we apply FA on the instance similarity baseline, BYOL, and achieve consistent improvement.

\begin{figure*}[t!]
\vspace{-0.3cm}
    \centering
    \subfigure[Basic FA framework]{
        \label{fig:framework1}
        \includegraphics[width=0.235\linewidth,height=0.22\linewidth]{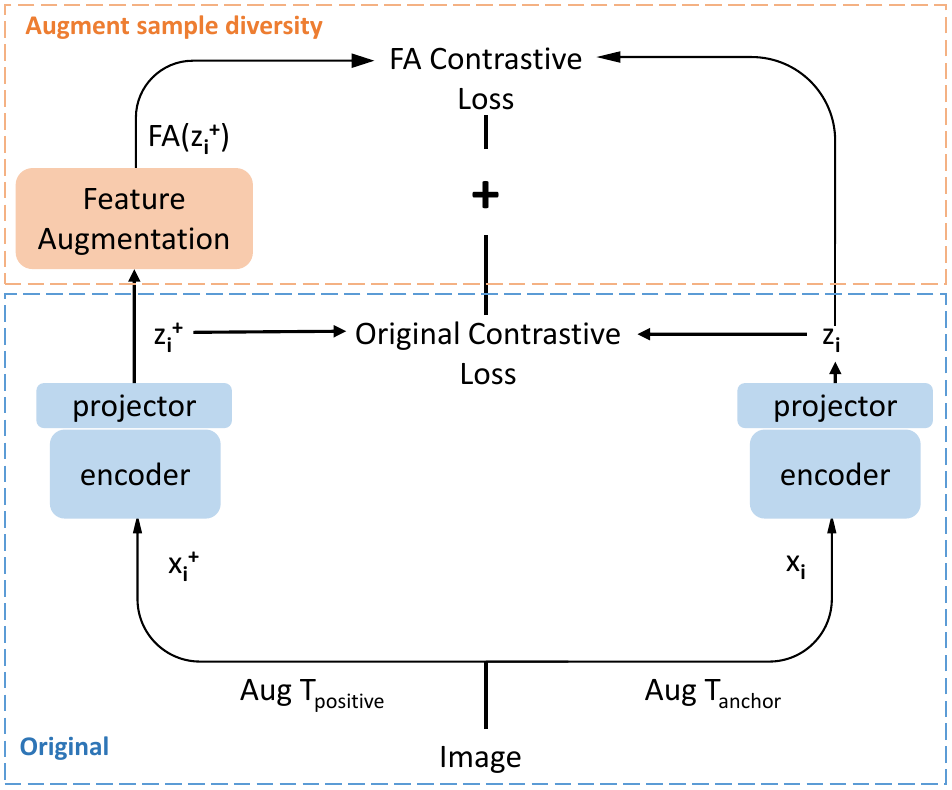}}
    \subfigure[Parallel-predictor-FA]{
        \label{fig:framework2}
        \includegraphics[width=0.24\linewidth,height=0.21\linewidth]{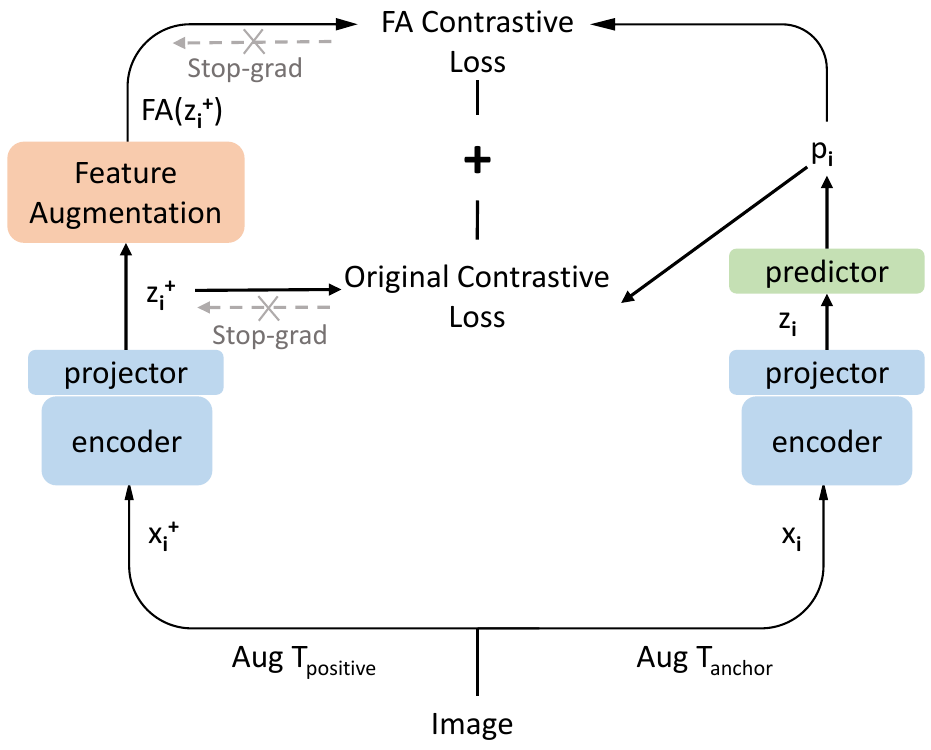}}
    \subfigure[Post-predictor-FA]{
        \label{fig:framework3}
        \includegraphics[width=0.24\linewidth,height=0.21\linewidth]{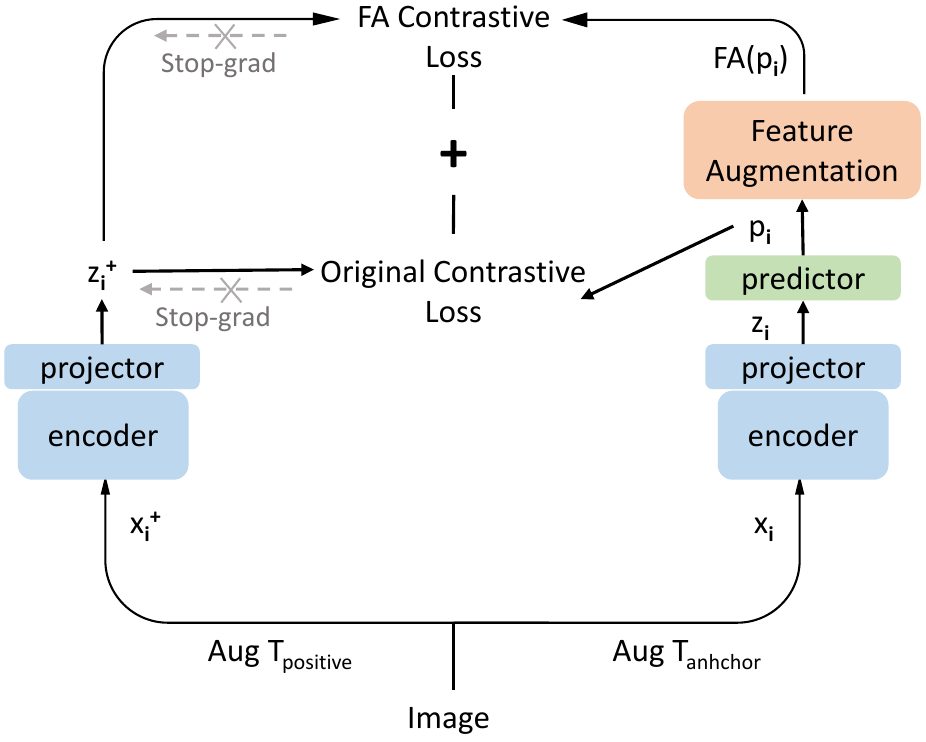}}
    \subfigure[Pre-predictor-FA]{
        \label{fig:framework4}
        \includegraphics[width=0.24\linewidth,height=0.21\linewidth]{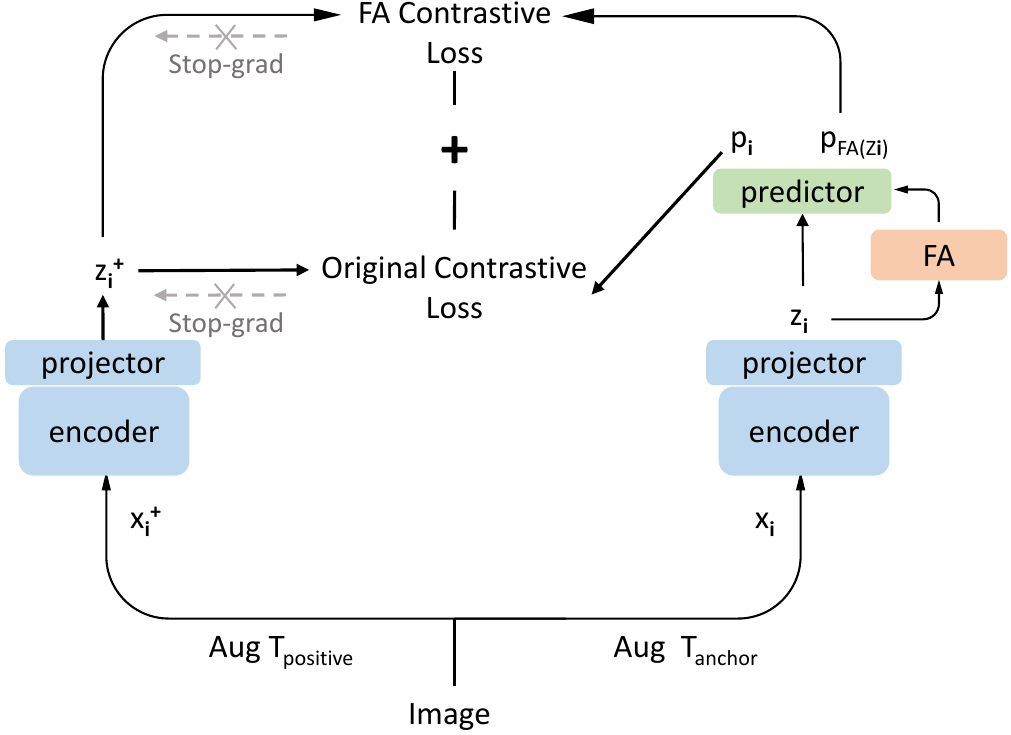}}
    \caption{Contrastive learning architectures with feature augmentation (FA). (a) is the basic framework. (b)-(d) extend (a) with additional predictors.}
    \label{fig:framework}
    \vspace{-0.2cm}
\end{figure*}

\subsection{The Basic Framework with Feature Augmentation}

The feature-level self-supervised works are based on different contrastive baselines. For example, the design of NNCLR and MSF relies on SimCLR and BYOL, with the heavy projector and predictor MLP head. But MoChi comes from MoCov1 with a light MLP head. Such a misalignment of architecture confuses the implementation of FA. To find a better architecture for FA, we start to build FA on the basic structure of SimCLR.

SimCLR is a basic structure of self-supervised contrastive learning, as shown in Fig. \ref{fig:framework1} (bottom part). Concretely, two pipelines of random data augmentation $T_{anchor}$ and $T_{positive}$ transform the same image into two views, which are forwarded by the same encoder+projector to get the positive feature pair ($z_i$, $z_i^+$), while the negatives come from all the other features in the current mini-batch. Notice that, following the setting of SimCLR, the encoder is ResNet-50 and the projector is a two-layer non-linear MLP \textit{without BN}.
Then we adopt the contrastive loss for the original data sample:
 \begin{equation}
\label{eqn:simclrloss}
\mathcal{L}_{i}^\text{original}(z_i, z_i^+) = -  \log{
	{
		\frac{\exp{(\ell_2(z_i)\cdot \ell_2(z_i^+)/\tau)}}
		{ \sum\limits_{k=1}^N\exp{(\ell_2(z_i)\cdot \ell_2(z_k^+) / \tau)}}
	}
}
\end{equation}
For one mini-batch with N samples, the overall loss is $\mathcal{L}^\text{original}= \frac{1}{N}\sum\limits_{i=1}^N \mathcal{L}^\text{original}_i$, which corresponds to the original contrastive loss.

The original contrastive loss only learns the variance brought by the prior data augmentation pipelines. Because of the shortage of views to construct more positive pairs, previous works provide practical solutions. For example, multi-crop \cite{caron2020unsupervised} augments additional small views to compare with two large views from the same image, to learn a more robust model. Though increasing the number of views from the input level is direct and powerful, the training memory/time/computation increases significantly with more views generated from the DA pipeline. 
Hence, to improve the sample diversity of the feature representation and step forward to multiple positive pairs with a small cost, we propose to use feature augmentation for the contrastive loss, as shown in Fig. \ref{fig:framework1}, 
After the feature augmentation, the new positive pair ($z_i$, $FA(z_i^+)$) will participate in the contrastive learning process. So we have:
\begin{equation}
\label{eqn:faloss}
\small
\mathcal{L}_{i}^\text{FA}(z_i, FA(z_i^+)) = -  \log{
	{
		\frac{\exp{(\ell_2(z_i)\cdot \ell_2(FA(z_i^+))/\tau)}}
		{ \sum\limits_{k=1}^N\exp{(\ell_2(z_i)\cdot \ell_2(FA(z_k^+)) / \tau)}}
	}
}
\end{equation}

The average loss of the mini-batch is $\mathcal{L}^\text{FA}= \frac{1}{N}\sum\limits_{i=1}^ N \mathcal{L}^\text{FA}_i$. 
The final loss of FA is the average of $\mathcal{L}^\text{FA}$ and $\mathcal{L}^\text{original}$ to combine the original and FA view. 
Note that $\ell_2$ normalization is applied in each contrastive loss, no matter for the original feature or the augmented feature. We ensure that the contrastive learning process happens in the unit sphere.

We have also attempted to construct the positive pair of two FA views, namely ($FA(z_i)$, $FA(z_i^+)$), which is too difficult for the model and even leads to non-convergence for some FA methods.
Therefore, we follow the positive pair setting of ($z_i$, $FA(z_i^+)$) to contrast one FA view with the original one. 

\begin{table*}[tb]
    \centering
    \small
    \caption{ImageNet-100 performance of various contrastive architectures with feature augmentation. \gbf{Green fonts} indicate increases over baselines, while \rbf{grey fonts} indicate declines. The Parallel-predictor-FA gets the largest performance gains.}
    \scalebox{0.85}{
    \begin{tabular}{c|cccccc}
        \toprule
Method & No FA    &  Mask   & NN  & NN noise & Batch noise  & Gaussian noise   \\
        \midrule

    Basic 	  &   77.1   & 77.4\gbf{+0.3}    &  75.9\rbf{-1.2}    & 77.4\gbf{+0.3}      & 77.0\rbf{-0.1}         & 77.3\gbf{+0.2}  \\
    
+strong proj	  &  77.8    & 78.4\gbf{+0.6}    &  77.7\rbf{-0.1}    & 78.0\gbf{+0.2}      & 78.2\gbf{+0.4}         & 78.8\gbf{+1.0}  \\
    \midrule
    Parallel-predictor-FA    & 78.4    & 78.3\rbf{-0.1}    &  \bf{79.6\gbf{+1.2}}    & \bf{79.3\gbf{+0.9}}      & \bf{79.2\gbf{+0.8}}         & \bf{78.9\gbf{+0.5}}  \\

    Post-predictor-FA   & 78.4      & 79.1\gbf{+0.7}    &  78.7\gbf{+0.3}    & 77.9\rbf{-0.5}      & 79.1\gbf{+0.7}         & 78.5\gbf{+0.1}  \\
    
    Pre-predictor-FA   & 78.4     & \bf{79.3\gbf{+0.9}}    &  77.0\rbf{-1.4}    & 78.5\gbf{+0.1}      & 78.8\gbf{+0.4}        & 78.6\gbf{+0.2}  \\
    \bottomrule
    \end{tabular}}
    \vspace{-0.5cm}
    \label{tab:framework}
\end{table*}

\subsection{Feature Augmentation Methods}

After the introduction of the basic FA framework, we now start to discuss the detailed FA methods. 
The motivation of recent self-contrastive works with feature space manipulation can be summarized as ``increasing the view variance to learn an invariant representation'' from the InfoMin principle \cite{tian2020makes}. For example, they leverage hard example mining \cite{kalantidis2020hard_mochi,zhu2021improving} and the nearest neighbor to replace the original feature \cite{dwibedi2021little,koohpayegani2021mean}. The purpose is to make the pre-training process harder and boost the model performance. 
However, our goal of FA is to augment more feature points close to the original feature. 
Using the augmented feature views, more contrastive learning processes can be organized for the pre-training and thus improve the robustness and generalization. 
Hence, we discard hard-example-mining methods which may be ``out-of-manifold'' and lead to non-convergence, like the extrapolation or large-scale mask. 

Instead, we integrate three feature augmentation methods from previous works:

{\textbf{(1) Nearest neighbor (NN):}}
NNCLR and MSF utilize the NN method to introduce more variance. It significantly boosts the performance of image classification. Following their setting, we employ a feature bank\cite{wu2018unsupervised} with the size $65536$ to store the $\ell_2$ normalization features from previous iterations. And we select the most similar $k$ features as the augmentations. 

\textbf{(2) Random dropout mask (Mask):}
SimCSE~\cite{gao2021simcse} utilizes random dropout masks on features to construct different positive embeddings. We apply random discrete dropout masks ($20\%$) on features to bring more robustness to the pre-trained model, i.e., the masked 256-d feature will retain $80\%$ original value while another $20\%$ will be zeros. The augmented $k$ features have $k$ various random masks. We have tried the mask rate of  $20\%$ and $50\%$ in our experiments, and the latter is harmful to the pre-training performance.

{\textbf{(3) Mixup interpolated noise:}}
It seems that mix-up is the most popular strategy in self-contrastive learning, no matter in the data-level \cite{verma2021towards,lee2021imix,shen2020mix} or the feature-level \cite{kalantidis2020hard_mochi,zhu2021improving}. 
Our motivation is to introduce noise and augment a few samples near the original features. Hence, we employ a mild mixup strategy  among them (i.e., DACL \cite{kalantidis2020hard_mochi}) but in the feature space by:
\begin{equation}
{f}_{aug} = \lambda\cdot f_{original} + (1 - \lambda)\cdot f_{noise}
\end{equation}
, where $\lambda \sim U(\alpha,1.0)$ is randomly sampled from a uniform distribution. We set $\alpha=0.85$ so that the augmented feature is less affected by the noise feature. 
We provide three choices of $f_{noise}$: \textbf{(i) NN noise}: the $k$ nearest neighbor of $f_{original}$; \textbf{(ii) Batch noise}: random selecting $k$ features from the same batch of $f_{original}$; \textbf{(iii) Gaussian noise}: random sampling $k$ gaussian noise from $ \mathcal{N}(0,0.2)$. 
We choose three kinds of these noises with different randomness to evaluate the efficacy of FA.

\subsubsection*{\textbf{Preliminary experimental evaluation of the basic framework with FA}}
Particularly, we focus on three unsolved open problems (section\ref{sec:intro}) and provide empirical solutions, to better equip FA for self-supervised contrastive models. 
We report the performance of the basic framework with FA on ImageNet-100 in Table \ref{tab:framework}. Please refer to the relative performance gain over the baseline.
In the first row of Table \ref{tab:framework} (i.e., \textit{Basic}), it appears that FA only brings a little performance to the contrastive pre-training, while some FA methods drop a lot (e.g., NN method). 
It is noteworthy that NNCLR applies the enhanced projector (three non-linear MLPs followed by batch normalization \cite{ioffe2015batch}). 
To align the experimental settings, we replace this 3-layer BN projector with a strong projector in the basic framework. As shown in Table \ref{tab:framework} (i.e., \textit{+strong proj}), using a strong projector indeed improves the performance of FA. We attribute that a strong projector with a large capacity arouses the efficacy of FA. Thus, we continue to use this strong projector to perform the remaining experiments.

\begin{table*}[htb!]
    \centering
    \small
    \caption{ImageNet-100 performance of each architecture with FA that does not use the stop-gradient operation.}
    \scalebox{0.85}{
    \begin{tabular}{c|cccccc}
        \toprule
    Method & ~~Baseline    &  ~Mask~   & ~~NN~~  & NN noise & Batch noise   & Gaussian noise    \\
        \midrule
    Parallel-predictor-FA     &  78.4    & 78.2\rbf{$\downarrow$}    &  77.7\rbf{$\downarrow$}    & 77.9\rbf{$\downarrow$}      & 78.3\rbf{$\downarrow$}         & 77.7\rbf{$\downarrow$}  \\
    Post-predictor-FA   &  78.4    & 77.4\rbf{$\downarrow$}    &  78.3\rbf{$\downarrow$}    & 78.3\rbf{$\downarrow$}      & 78.6\gbf{$\uparrow$}         & 78.0\rbf{$\downarrow$}  \\
    Pre-predictor-FA   &  78.4    &  77.8\rbf{$\downarrow$} & 77.9\rbf{$\downarrow$}  &    78.5\gbf{$\uparrow$} &  78.4\rbf{$\downarrow$}           & 78.0\rbf{$\downarrow$}  \\
    \bottomrule
    \end{tabular}}
    \label{tab:stopgrdient}
\end{table*}

\begin{table*}[htb!]
    \centering
    \small
    \vspace{-0.1cm}
    \caption{ ImageNet-100 performance of FA when applying different settings of data augmentation pipeline }
    \scalebox{0.85}{
    \begin{tabular}{c|cccccc} 
        \toprule
        Method   & ~Baseline &  ~Mask~    & ~~NN~~  & NN noise    & Batch noise   & Gaussian noise  \\
        \midrule
            SymmWeakAug & 74.4 & 75.6\gbf{+1.2}  & 77.6\gbf{+2.2}   & 76.1\gbf{+1.7}  & 75.4\gbf{+1.0}  &75.9\gbf{+0.5}    \\
        SymmStrongAug   &78.0  & 78.4\gbf{+0.4} & 78.9\gbf{+0.9}  & 78.3\gbf{+0.3} & 78.5\gbf{+0.5}   & 78.5\gbf{+0.5}  \\

    AsymmStrongAug    &77.8 & 78.5\gbf{+0.7}  &79.4\gbf{+1.6}   &79.1\gbf{+1.3}  &79.4\gbf{+1.6}  &78.8\gbf{+1.0}  \\
        \bottomrule
    \end{tabular}}
\vspace{-0.3cm}
    \label{tab:dafa}
\end{table*}

\subsection{Feature Augmentation Architectures with Predictors}

We extend the basic framework (Fig. \ref{fig:framework1}) to three additional network layouts as shown in Fig. \ref{fig:framework}, by equipping FA with predictor modules. This is inspired by the utilization of the predictor after the projected features in BYOL \cite{grill2020bootstrap}. 
BYOL is a kind of instance similarity learning method without discriminating the negative samples, where the model tends to collapse during pre-training. Directly measuring the similarity of ($z_i$, $z_i^+$) in projector space can lead to model collapse if they are constant across views. 
The predictor further introduces more transformations by a parameterized non-linear MLP layer. Thus, the predictor is applied asymmetrically on one single view $z_i$ and gets the prediction $p_i$.
As a result, the learning objective becomes more challenging when maximizing the similarity between the views in different feature spaces ($p_i$, $z_i^+$) rather than views in the same feature space ($z_i$, $z_i^+$).
Besides, MSF and NNCLR also apply the predictor after the projected features with feature manipulations. Such cross-level positive pair and contrasting improve the view variance for the pre-training process.  

Therefore, we equip FA with the predictor for contrasting views in various feature spaces
We can regard FA and the predictor module as two kinds of transformations: the predictor is a parametric transformation by a learnable non-linear MLP head, and FA augments the samples close to the original feature.  
Naturally, we ask: where to add the predictor and FA, by applying the two transformation strategies on both projected views or a single view? Regarding this, we have designed three architectures as shown in Fig. \ref{fig:framework2}, \ref{fig:framework3} and \ref{fig:framework4}:

{\textbf{(1) Parallel-predictor-FA:}}
The first idea is to separate FA and the predictor into two views independently, which looks like parallelly applying two kinds of transformation methods on two branches. As shown in Fig. \ref{fig:framework2}, we apply FA on the positive view and projector on the anchor view to construct the cross-level positive pair ($p_i$, $FA(z_i^+)$). Compared with the basic framework ($z_i$, $FA(z_i^+)$), the predictor brings the transformation from the MLP layer with a more challenging robust contrastive process. The final augmented loss is the average of the cross-level loss $\mathcal{L}^\text{FA}(p_i, FA(z_i^+))$ and $\mathcal{L}^\text{original}(p_i, z_i^+)$, which combines the original and FA view. In a word, the Parallel-predictor-FA architecture contrasts the predicted anchor view with the FA view and the original one.

{\textbf{(2) Post-predictor-FA:}}
In this model, we apply the predictor and FA on the same side. As shown in Fig. \ref{fig:framework3}, we use FA after the predictor. This design applies two transformations (the predictor followed by FA) on the anchor view, and it contrasts with the positive projected view ($FA(p_i)$, $z_i^+$). The final augmented loss is the average of the cross-level loss $\mathcal{L}^\text{FA}(FA(p_i), z_i^+)$ and $\mathcal{L}^\text{original}(p_i, z_i^+)$.

{\textbf{(3) Pre-predictor-FA:}}
Another form of performing both FA and predictor on the same side is to first apply FA and then predictor on the anchor view. The augmented cross-level positive pair ($p_{FA(z_i)}$, $z_i^+$) also contrasts the double-transformed anchor view with the positive projected view. 
The final augmented loss is the average of the cross-level loss $\mathcal{L}^\text{FA}(p_{FA(z_i)}, z_i^+)$ and $\mathcal{L}^\text{original}(p_i, z_i^+)$.

\subsubsection*{\textbf{Experimental evaluation of different FA architectures}}
We perform experiments on ImageNet-100 to evaluate which architecture could better utilize the predictor and enhance the efficacy of FA. Please refer to the relative performance gains over the baseline.
As shown in Table \ref{tab:framework}, applying FA and projector parallel on two views can provide the maximum benefit on most FA methods except the mask. Actually, the mask method can benefit from single-side architecture($79.1\%$ and $79.3\%$ for Post/Pre-predictor-FA architecture).
When only one view experiences too many transformations, the performance improvement will not be comparable with the Parallel-predictor-FA architecture and even drops a lot for the NN method.
To prevent the scenario of out-of-manifold, we discard Post/Pre-predictor-FA architecture and adopt the Parallel-predictor-FA architecture with good performance. 
Please notice that, when the baseline is strong (i.e., with higher performance), it generally becomes harder to gain further performance improvement. Hence, the Parallel-predictor-FA architecture indeed enhances the efficacy of FA. 
Besides, such Parallel-predictor-FA architecture has been utilized in existing self-contrastive pre-training works, such as OBoW \cite{gidaris2021obow}, DINO\cite{caron2021emerging}, BYOL, MSF and NNCLR. However, these methods have not validated that this architecture is suitable for FA.

\subsubsection*{\textbf{The importance of stop-gradient}}
Besides the predictor, BYOL also applies the momentum encoder and stop-gradient to make $z_i^+$ stable and more challenging for the instance similarity task. 
However, our framework is a single-encoder model. We could not apply the EMA (i.e., exponential moving average, the updating strategy of the momentum encoder) to stabilize the positive projected view $z_i^+$. Therefore we only employ the stop-gradient to control the objective loss.
The effect of stop-gradient can be understood as an optimization strategy to prevent over-fitting. The non-parametric transformation of FA might be overfitted during the pre-training, so we apply the stop-gradient to the projected features (blocking the gradient flow of $z_i^+$ or $FA(z_i^+)$) to prevent it from the model collapse.
As shown in Table \ref{tab:stopgrdient}, when we remove the stop-gradient of three predictor-based architectures, the performance gains of FA become almost negative. Without stop-gradient, the benefits of FA strategy start to vanish, and even harm the final performance. Therefore, we conclude that stop-gradient is a key ingredient for the success of feature augmentation.

\subsection{Exploring the Relationship with Data Augmentation} \label{sec:dafa}
Even though previous works have extensively discussed the data augmentation\cite{koohpayegani2021mean,chen2020simple,grill2020bootstrap}, the relationship between DA and FA is still unclear.
For example, although DA and FA are two independent components, can FA make up a deficiency of DA? Moreover, what kind of DA pipeline setting is suitable for an effective feature augmentation?
In this section, we discuss their relationships to find out a better data augmentation setting for FA. Based on the Parallel-predictor-FA architecture, we design three DA settings:

{\textbf{(1) Symmetric weak augmentation (SymmWeakAug):}}
In this setting, we remove two DA strategies, i.e., Gaussian blurring \cite{chen2020simple} and solarization \cite{grill2020bootstrap}. These are relatively violent means compared with the color or grey change. Notably, the parameter setting of DA is symmetric.

{\textbf{(2) Symmetric strong augmentation (SymmStrongAug):}}
In this setting, we adopt all the DA strategies proposed in BYOL with different parameters. Similarly, the parameter setting of DA is symmetric, namely $T_{anchor}=T_{positive}$.

{\textbf{(3) Asymmetric strong augmentation (AsymmStrongAug):}}
This setting also employs all the DA strategies in BYOL. But the parameter setting of the DA pipeline is asymmetric, $T_{anchor}\not=T_{positive}$. We keep all strategies in one pipeline while removing Gaussian blurring and solarization from another pipeline.

\begin{figure}[tb!]
    \begin{center}
        \includegraphics[width=0.85\linewidth]{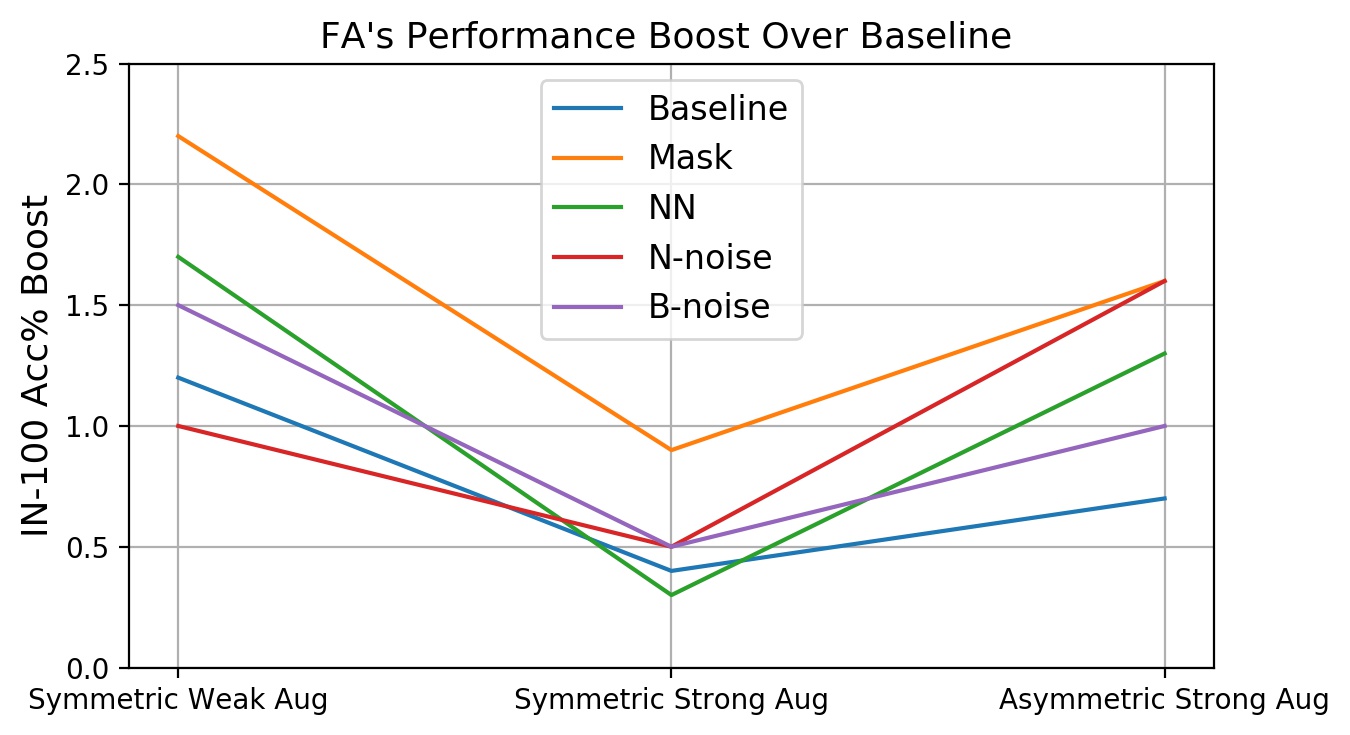}
    \end{center}
    \vspace{-0.4cm}
    \caption{The performance boost of feature augmentation (FA) over baselines when applying different settings of data augmentation (DA). FA makes up a deficiency for DA, and asymmetric DA setting is suitable for FA.}
    \label{fig:dafaboost}
    \vspace{-0.4cm}
\end{figure}

The detailed evaluation results are listed in Table \ref{tab:dafa}. For a better comparison, we also draw the performance boost of FA over the baseline in Fig. \ref{fig:dafa}. 
From SymmStrongAug to SymmWeakAug, the relative performance boost of FA has been improved a lot. This indicates that FA can better improve DA effectively. The SymmWeakAug+NN setting ($77.6\%$) can even catch up with the performance of the baseline with strong augmentation ($78.0\%$). MSF and NNCLR also demonstrate that their feature space manipulation is helpful in the weak-DA setting.  

In addition, from SymmStrongAug to AsymmStrongAug, the FA's relative performance has been significantly boosted across all FA methods. This comparison demonstrates that the asymmetric parameter setting of the DA pipeline is more appropriate for feature augmentation. Hence, we employ BYOL's DA parameters as the default setting for our FA experiments except those particularly mentioned. The asymmetric DA could introduce more variance for the positive pair during pre-training. By combining asymmetric DA, FA and predictor, we can construct a better contrastive model.

\subsection{Upgrading BYOL with Feature Augmentation}\label{sec:byol}

After previous analysis and empirical evaluation, we adopt the Parallel-predictor-FA architecture, stop-gradient strategy and the data augmentation setting for BYOL \cite{grill2020bootstrap}. 
Next, we evaluate the efficacy of FA on the instance similarity framework, BYOL. The Parallel-predictor-FA architecture is very similar to BYOL's, which directly adds FA on the top of the momentum encoder and applies cross-level contrast.
As shown in Table \ref{tab:byol}, all FA methods bring consistent improvements based on the BYOL baseline. We utilize the original structure of BYOL without a strong projector. This performance boost indicates that the Parallel-predictor-FA architecture and stop-gradient are practical for feature augmentation. 

\begin{table}[tb!]
    \centering
    \small
    \caption{ImageNet-100 performance on BYOL with feature augmentation. The baseline's accuracy is $77.4\%$. FA brings improvement for BYOL in various degrees. Augmenting more samples and freeing the loss further improves the accuracy}
    \setlength\tabcolsep{4pt} 
    \scalebox{0.82}{
    \begin{tabular}{c|ccccc} 
        \toprule
        Method   &  ~Mask~    & ~~NN~~  & NN noise    & Batch noise   & Gaussian noise    \\
        \midrule
            Aug 1 & 77.9\gbf{+0.5}    & 78.7\gbf{+1.3}  & 79.0\gbf{+1.6}  & 78.8\gbf{+1.4}   & 78.3\gbf{+0.9}   \\
            Aug 4      & 78.1\gbf{+0.7}  & 77.7\gbf{+0.3} & 79.0\gbf{+1.6} & 78.4\gbf{+1.0}  &  78.9\gbf{+1.5} \\
        Aug 1 free     & 79.6\gbf{+2.2}  & \textbf{79.5\gbf{+2.1}} & \textbf{79.5\gbf{+2.1}}  & 79.2\gbf{+1.8} & \textbf{79.1\gbf{+1.7}}  \\
    Aug 4 free       & \textbf{79.8\gbf{+2.4}}  & 79.1\gbf{+1.7}  & 78.7\gbf{+1.3} & \textbf{79.4\gbf{+2.0}} & 79.1\gbf{+1.7}  \\
        \bottomrule
    \end{tabular}}
    \label{tab:byol}
    \vspace{-0.3cm}
\end{table}

\textbf{Augment more samples and free the loss:}
With the help of FA, we can obtain more samples similar to the original feature point. As a result, we augment more samples to see whether FA can provide more diverse positives in the feature space. We try to augment $4$ samples in the feature space. The final loss is the average of the original loss and four augmented losses.
As listed in the second row of Table \ref{tab:byol}, augmenting $4$ samples does not gain better performance than only augmenting one sample. We conjecture that the average operation in the final loss may limit the effect of FA. Thus, we free the loss (i.e., remove the average operation in the final loss) and use the $2\times$ or $5\times$ of the previous loss when augmenting $1$ or $4$ samples.  
The final two rows in Table \ref{tab:byol} show the consistent improvement of free loss over the baseline. The free-loss strategy gains higher accuracy than non-free loss, which suggests FA has great potential to boost a pre-trained model. 

\section{Experiments: Pre-training \& Transferring}

\begin{table*}[htb!]
    \centering
    \small
    \vspace{-0.3cm}
    \begin{center}
        \caption{Linear evaluation on ImageNet-1k. FA consistently improves the baseline for instance discrimination and instance similarity}
    \scalebox{0.9}{
    \begin{tabular}{lccc|lccc}
        \toprule
        Instance   & Batch & ~Epochs~ & Linear& ~Instance & Batch&~Aug~& Linear  \\
        
     Discrimination  & Size &  & Acc\% &  ~Similarity&  &~num~& Acc\%  \\
        \midrule
        \multicolumn{4}{l}{\textit{Official reported results}}  \\
        NNCLR\cite{dwibedi2021little} & 4096 & 100/200 & 69.4/70.7 &  &   &&  \\
         \midrule
        Baseline & 512 & 100/200 & ~66.0/68.4~ &~BYOL$^{\dagger}$ &512 & - & 69.1 \\
        + Mask & 512 & 100/200 & ~67.2/69.4~ &~+Mask&512 & 1/4 & 69.5/70.0 \\
        + NN & 512 & 100/200 & ~68.6/70.9~ &~+NN&512 & 1/4 & 71.4/70.9 \\
        + NN noise & 512 & 100/200 & ~67.3/69.3~ &~+NN noise&512& 1/4 & 70.0/70.8 \\
        + Batch noise & 512 & 100/200 & ~67.4/69.5~ &~+Batch noise&512 & 1/4 & 69.8/70.2 \\
        + Gaussian noise & 512 & 100/200 & ~67.4/69.4~ &~+Gaussian noise&512 & 1/4  & 69.9/70.1 \\
        \bottomrule
        \end{tabular}}
        \end{center}
    \vspace{-0.3cm}
\label{tab:imagenet}
\end{table*}

\begin{table}[htb!]
    \begin{center}
    \vspace{-0.5cm}
    \caption{Transferring to Object Detection on PASCAL VOC (IN-1k stands for ImageNet-1k).}
    \setlength\tabcolsep{3pt}
    \scalebox{0.8}{
        \begin{tabular}{lccccc|ccccc}
            \toprule
            Method  & Epochs & IN-1k & AP & $\text{AP}_{50}$  & $\text{AP}_{75}$ & Epochs & IN-1k & AP & $\text{AP}_{50}$  & $\text{AP}_{75}$  \\
            \midrule
            Baseline  & 100 & 66.0  & 52.6 & 80.7  & 57.5  & 200 & 68.4 & 53.7 & 81.2  & 59.5  \\
           
            
            + Mask  & 100 & 67.2  & 54.4 & 81.6   & 60.4 & 200 &  69.4 & 55.1 & 82.0   & 61.9 \\

            + NN & 100 &  \bf{68.6} & \rrbf{52.5} & \rrbf{80.7}   & \rrbf{57.1}  & 200 &     \bf{70.9} & \rrbf{53.3} & \rrbf{81.2}  & \rrbf{58.5} \\           


           + NN noise  & 100 &  67.3 & 54.6 & 81.6   & 60.2  & 200 & 69.3  & 55.3 & 82.3   & 61.9 \\
            
            + Batch noise  & 100 & 67.4  & 54.4 & 81.4   & 60.3 & 200 &  69.5 & 55.3 & 82.1   & 60.6 \\
            
            + Gaussian noise  & 100 & 67.4  & 54.8 & 81.3   & 60.5  & 200 & 69.4  & 55.3 & 81.8   & 61.4 \\            
            \bottomrule
        \end{tabular}
    }
    \end{center}
    \vspace{-0.6cm}
    \label{tab:detection}
\end{table}

In this section, we report the results of self-supervised contrastive learning (on ImageNet2012 \cite{imagenet2015} with 1,000 classes) over both the instance discrimination (i.e., cross-entropy-based) \cite{chen2020simple} and instance similarity (i.e., prediction-based) \cite{grill2020bootstrap} paradigms, and transfer learning (object detection on PASCAL VOC \cite{everingham2010pascal}) to evaluate the generalization of FA. Pytorch \cite{paszke2019pytorch} and the Pytorch-Lightning library \cite{Falcon_PyTorch_Lightning_2019} are used for all experiments.

\subsection{Self-supervised Pre-training on ImageNet}\label{sec:in1k-exp}

\textbf{Instance discrimination:}
We first evaluate the efficacy of FA based on instance discrimination contrastive learning using the Parallel-predictor-FA architecture. Following the same setting of NNCLR \cite{dwibedi2021little}, we use ResNet-50 as the backbone, with a strong projector and a 3-layer-MLP followed by BN and ReLU activations (no ReLU in the last layer of the projector).  
The size of the projector is $[2048,2048,256]$. The predictor is a 2-layer-MLP with the size of $[4096,256]$ and no BN or ReLU in the last layer.
We apply the SGD optimizer (lr=0.4 with cosine scheduler, momentum=0.9, weight decay=$10^{-5}$,  cosine warm-up $10$ epochs) to learn the pre-trained model for 100 or 200 epochs. 
The batch size is 512 due to the limitation of computation capability and we use accumulating gradient to make up the gap to the large batch size (i.e., $4096$ in NNCLR). 
LARS \cite{you2017large} is also applied with the trust coefficient $0.02$. As aforementioned in Section \ref{sec:dafa}, we directly adopt the setting of the DA pipeline in BYOL~\cite{grill2020bootstrap}. The temperature $\tau$ in contrastive loss is $0.2$. Notice that Sync BN and gradient clip are applied. We augment only one sample. No free loss is utilized in the instance discrimination model.

\textbf{Instance similarity:}
We apply the SGD optimizer (lr=0.45 with cosine scheduler, momentum=0.9, weight decay=$10^{-6}$, cosine warm-up $10$ epochs) and train for 100 epochs. The LARS's trust coefficient is set to $0.01$ and here gradient clip is unnecessary. We augment one and four samples and employ the free loss for this model. The momentum parameter starts from 0.99 to 1.0. Other settings keep the same with Instance Discrimination.

\subsection{Evaluation on ImageNet-1k}
We evaluate the pre-trained representation by training a linear layer on top of the frozen backbone. It is trained with the SGD optimizer (epochs=60, batchsize=256, weight decay=0, momentum=0.9, lr=1.0 / 0.2 for instance discrimination / similarity, decaying at $[40,50,55]th$ epochs). The results on ImageNet-1k are shown in Table \ref{tab:imagenet}.

For instance discrimination, FA methods bring consistent improvements over the baseline model, no matter in 100/200 training epochs as shown in Table \ref{tab:imagenet}. These improvements indicate that feature augmentation is also effective when pre-training on the large-scale dataset. Among the FA methods, NN brings the biggest improvement, about $2.5\%$ over the baseline model on 100/200 epochs setting. 
We also compare the baseline+NN model with NNCLR which is the state-of-the-art self-contrastive model. The baseline+NN cannot catch up with NNCLR in the 100-epoch setting, which can be attributed to our smaller batch size. Note that due to computation limitations, we only use 512 batch-size with gradient accumulation. Because the instance discrimination method uses the mini-batch as negative samples for contrastive learning, a smaller batch size means less data being explored.
However, when training longer to 200 epochs, the performance of baseline+NN catches up with NNCLR ($70.9\%$ \emph{v.s.} $70.8\%$). This indicates that feature augmentation can provide sample diversity and mitigate the limitation of data lack during pre-training.

For instance similarity, FA methods produce various degrees of performance boosts over the BYOL model as Table \ref{tab:imagenet} shows. When augmenting $1$ sample in the feature space, the random Mask only improves marginally ($69.5\%$ \emph{v.s.} $69.1\%$, $0.4\%$ improved) and the single NN significantly enhances the BYOL ($71.4\%$ \emph{v.s.} $69.1\%$, $2.3\%$ improved).
When the augment number increases to $4$, the performance of FA-NN degrades slightly while other FA methods get higher accuracy than augmenting only one sample.
The performance drop of NN is consistent with NNCLR\cite{dwibedi2021little}, the best result usually comes from the most similar feature.
Overall, FA indeed boosts the BYOL regardless of 1 or 4 samples in feature space. We use free loss in this experiment.

\subsection{Transfer Learning on Object Detection}
To evaluate the generalization of FA, we apply Faster R-CNN \cite{ren2015faster} for downstream object detection. We follow the procedure and default parameters adopted in \cite{he2020momentum} based on Detectron2 \cite{wu2019detectron2}. The pre-training parameters are fine-tuned on PASCAL VOC \cite{everingham2010pascal_VOC} trainval07+12 and evaluated at test07 set. The results are reported in Table \ref{tab:detection}. 
As we can see, FA methods (Mask and three Mixup interpolated noise) consistently improve over the baseline when transferring to the detection task. Also, we can conclude that these four FA methods can help the baseline model generalize better on the downstream detection task.
However, the NN methods slightly hurt the baseline, even if NN gets the highest results on ImageNet-1k linear evaluation.
We argue that the NN methods concentrate on finding the most semantically similar feature, which could overfit during the pre-training process on the ImageNet-1k dataset. Hence, NN does not improve the generalization although achieves a good performance on ImageNet-1k linear evaluation. 
In summary, FA can tackle the task-bias problem and generalize well thanks to less data bias.

\section{Conclusion}
We introduce and empirically evaluate the efficacy of feature augmentation for self-supervised contrastive learning. We reconsider feature augmentation as a basic strategy to improve sample diversity, and we solve three open problems ignored by previous feature manipulation works.
By integrating the proposed FA methods, we verify the proper architecture, ingredient component (stop-gradient) and the relationship with data augmentation, for this new strategy. 
Experiments on baselines of the instance discrimination and instance similarity paradigms show that FA consistently brings performance improvement and good generalization on downstream tasks.

\bibliographystyle{IEEEtran}
\bibliography{IEEEabrv,ijcnn24_refs}

\end{document}